\documentclass[10pt]{article}
\usepackage{examples,a4wide}

\newcommand{\emth}[1]{\ensuremath{#1}}

\newcommand{\ben}{\begin{enumerate}}
\newcommand{\een}{\end{enumerate}}
\newcommand{\bqt}{\begin{quote}}
\newcommand{\eqt}{\end{quote}}
\newcommand{\bit}{\begin{itemize}}
\newcommand{\eit}{\end{itemize}}
\newcommand{\bt}{\begin{tabbing}}
\newcommand{\et}{\end{tabbing}}
\newcommand{\btr}{\begin{tabular}{llll}}
\newcommand{\etr}{\end{tabular}}
\newcommand{\bd}{\begin{description}}
\newcommand{\ed}{\end{description}}

            % ...
\newcommand{\gG}    {\emth{\gamma \;}}            %
\newcommand{\gD}    {\emth{\delta \;}}           %
           % ...
\newcommand{\cC}    {\emth{\mathcal{C} \;}}           %
\newcommand{\cL}    {\emth{\mathcal{L} \;}}           %
\newcommand{\cP}    {\emth{\mathcal{P} \;}}           %
\newcommand{\cR}    {\emth{\mathcal{R} \;}}           %
\newcommand{\cT}    {\emth{\mathcal{T} }}           %

\newcommand{\aWHILE} {\emth{{{\sf \underline{while}}\;}}}
\newcommand{\aOR}    {\emth{{{\sf \underline{or}}\;}}}
\newcommand{\aAND}    {\emth{{{\sf \underline{and}}\;}}}

\newcommand{\aDO}    {\emth{{{\sf \underline{do}}\;}}}
\newcommand{\aIF}    {\emth{{{\sf \underline{if}}\;}}}
\newcommand{\aTHEN}  {\emth{{{\sf \underline{then}}\;}}}
\newcommand{\aELSE}  {\emth{{{\sf \underline{else}}\;}}}

\newcommand{\aRETURN} {\emth{{{\sf \underline{return}}\;}}}
\newcommand{\aCOMMENT}[1] {\emth{\triangleright} {\small #1}}

\newcommand{\ALPHA}[1] {\emth{\alpha_{#1}}}
\newcommand{\BETA}[1]  {\emth{\beta_{#1}}}
\newcommand{\SIGMA}[1] {\emth{\sigma_{#1}}}
\newcommand{\opRD} {\emth{\swarrow}}          
\newcommand{\opLD}  {\emth{\nearrow}}          
\newcommand{\calcAB}  {{\sf AB}}                
\newcommand{\calcF}  {{\sf F}}                  
\newcommand{\calcL}  {{\sf L}}                  
\newcommand{\calcLP}  {{\sf LP}}
\newcommand{\calcNL}  {{\sf NL}}
\newcommand{\calcLPE}  {{\sf LPE}}
\newcommand{\calcLPC}  {{\sf LPC}}
\newcommand{\calcLPCE}  {{\sf LPCE}}

\newcommand{\OA}    {\emth{\langle}}          
\newcommand{\CA}    {\emth{\rangle}}    
\newcommand{\PO}  {\emth{\sqsubseteq \;}}     
\newcommand{\ID}  {\textnormal{\texttt{1}}}   
\newcommand{\EL}    {\emth{\in \;}}           
\newcommand{\SE}[1] {\{#1\}}             

\newcommand{\ED}   {\emth{\buildrel \rm def \over = \;}}
\newcommand{\NE}   {\emth{\lnot}}             
\newcommand{\IM}   {\emth{\Rightarrow\;}} 
\newcommand{\MI}   {\emth{\Leftarrow\;}} 
\newcommand{\BU}   {\emth{\bullet \;}}        
\newcommand{\CI}   {\emth{\circ \;}}          
\newcommand{\BS}   {\emth{\backslash}}          
\newcommand{\EN}   {\emth{\vdash \;}}           
\newcommand{\ENI}[1]   {\emth{\vdash_{#1} \;}}  
\newcommand{\BB}   {\rule{2mm}{2mm}}     
\newcommand{\QED}  {\BB}                 

\newcommand{\RED}[1]   {{\em Definition \ref{#1}}}
\newcommand{\REF}[1]   {{\em Figure \ref{#1}}}

\newtheorem{definic1}{Definition}

\newcommand{\DEF}[2]
 {
  \begin{definic1}
        \label{#2}
        #1
  \end{definic1}
 }

\newtheorem{proposition1}{Proposition}

\newcommand{\FOR}[2]
 {
  \begin{eqnarray}
        #2 \label{#1}
  \end{eqnarray}
 }

\newtheorem{proof}{Proof}

\newenvironment{pseudocode}[2]
{ \vspace{-1.3ex}
  \begin{tabbing}
    \mbox{{\sf #1}} \\
    99 \= \kill                  
    #2}
  {
  \end{tabbing}}

\title{A Labelled Analytic Theorem Proving Environment
 for Categorial Grammar
}

\author{
 Saturnino F. Luz-Filho\thanks{Supported by CNPq
   Brazilian Research Council Research Studentship No. 200210/93-9} \\
   Patrick Sturt\thanks{Supported by ESRC  Research Studentship No.
R00429334338}\\
 \\
{\tt \{luz,sturt\}@cogsci.ed.ac.uk} \\
Centre for Cognitive Science \\
2 Buccleuch Place
Edinburgh  EH8 9LW \\
 Scotland, UK\\}

\date{}

\begin{document}

\maketitle

\begin{abstract}

We present a system for the investigation of computational
properties of categorial grammar parsing based on a labelled analytic
tableaux theorem prover. This proof method allows us to take  a modular
approach, in which the
basic grammar can be kept constant, while a
range of categorial calculi can be captured by assigning different properties
to the
labelling algebra. The theorem proving strategy is particularly well
suited to the treatment of categorial grammar, because it allows us to
distribute the computational cost between the algorithm which deals
with the grammatical types and the algebraic checker which constrains
the derivation.

\end{abstract}

\section{Background}
\label{sec:Background}

A current trend in logic is to attempt to incorporate semantic
information into the domain of
deduction, \cite{gabbay94}, \cite{barwiseetal95}.
An area for which  this strategy is particularly useful is the problem of
categorial
grammar parsing. The categorial grammar research programme requires
the use of a range of logical calculi for linguistic
description. Some researchers have considered labelled deduction as a
tool for implementing categorial parsers \cite{moortgat92},
\cite{morrill95}, and this paper can be seen as a new contribution to
this field.

In this paper we aim for a modular approach, in which the basic
grammar is kept constant, while different calculi can be implemented
and experimented with by constraining the derivations produced by the
theorem prover.  At present, our system covers the classical Lambek
Calculus, \calcL{}, as well as the non-associative Lambek calculus
\calcNL{}, \cite{lambek61}, and variants such as Van Benthem's
\cite{vanbenthem88} \calcLP{}, \calcLPC{}, \calcLPE{} and \calcLPCE{},
and their non-associative counterparts. The system is based on
labelled analytic deduction, particularly on the LKE method, developed
by D'Agostino and Gabbay \cite{dagostino94}.  LKE is similar to a
Smullyan-style tableau system, in which the derivations obey the
sub-formula principle, but it improves on efficiency by restricting
the number of branching rules to just one. Different categorial logics
are handled by assigning different properties to the labelling
algebra, while the basic syntactic apparatus remains the same.  This
allows the user to experiment with various linguistic properties
without having in principle to modify the grammar itself.

The basic structure of the paper is as follows. In section
\ref{sec:Family}, we introduce the family of categorial calculi, and
discuss some of the linguistic arguments which have been put
forward in the literature with regard to these calculi.
In section \ref{sec:Framework}, we introduce the logical
apparatus on which the system is based, describe the algorithm and
prove some of the  properties mentioned in section \ref{sec:Family}
within this framework. We also show how different grammars can be
characterised and present a worked example.
In section \ref{sec:Comparison} the system is compared with other
strategies for  dealing with  multiple categorial logics, such as
hybrid formalisms and unification-based Gentzen-style deduction. In
this section, we also suggest some ways to improve the efficiency of
the system, and strategies  for dealing with the complexity of
labelled unification.

\subsection{A Family of Categorial Calculi and Their Linguistic Applications}
\label{sec:Family}

Categorial Grammars can be formalised in terms of a hierarchy of well
understood and mathematically transparent logics, which yield as
theorems  a range of combinatorial operations. However the precise
 nature of the
combinatorial power required for an adequate characterisation of
natural language is still very much a matter of debate. For this
reason, it is desirable to have a means of systematically testing the
linguistic consequences of adopting various calculi. In this section
we give an overview of the linguistic applications of some of the
calculi in the hierarchy, with a view towards motivating the
usefulness of a generic categorial theorem prover as a tool for
linguistic study.

 The combinatorial
possibilities of expressions in general can be characterised in terms
of  {\em reduction laws}. In {\em R1}-{\em R6} below, we give some  reduction
laws discussed in \cite{moortgat88}, which have been found to be
linguistically useful.\footnote{We adopt the Lambek notation, in which
  X$/$Y is a function which ``takes'' a Y to its right to yield an X,
  and Y$\backslash$X is a function which ``takes'' a  Y to its left to
  yield  an X.}.

{\small
\begin{center}
\begin{tabular}[ht]{llllll}
{\em R1:} & {\em Application} & {\em R2:} & {\em Composition} & {\em
  R3:} & {\em Associativity} \\
   &  X$/$Y, Y \EN X & & X$/$Y, Y$/$Z \EN X$/$Z &
   & (Z$\backslash$X)$/$Y  \EN Z$\backslash$(X$/$Y) \\
   & Y, Y$\backslash$X \EN X & & Z$\backslash$Y,
   Y$\backslash$X  \EN Z$\backslash$X & &
   Z$\backslash$(X$/$Y) \EN (Z$\backslash$X)$/$Y \\ & & & & & \\
{\em R4:} & {\em Lifting} & {\em R5:} & {\em Division (main functor)}
&  {\em R6:} & {\em Division (sub. functor)} \\
& X \EN Y$/$(X$\backslash$Y) & & X$/$Y \EN
(X$/$Z)$/$(Y$/$Z) & & X$/$Y \EN (Z$/$X)$\backslash$(Z$/$Y)\\
& X \EN (Y$/$X$)\backslash$Y & & Y$\backslash$X
\EN (Z$\backslash$Y)$\backslash$(Z$\backslash$X) & &
Y$\backslash$X \EN (Y$\backslash$Z)$/$(X$\backslash$Z)\\
\end{tabular}
\end{center}}

It is possible to define a hierarchy of logical calculi, each of which
admits one or more of {\em R1}-{\em R6} as theorems; from the purely
applicative  calculus {\sf AB}, of Ajdukiewicz and Bar-Hillel,
\cite{ajduk35}, which supports only {\em R1}, to the full Lambek
calculus {\sf L}, which supports all the above laws. Calculi
intermediate in power between  {\sf AB} and {\sf L} have been
explored (e.g. Dependency Categorial Grammar \cite{pickering93}), as
well as stronger calculi which extend the power of {\sf L} through
the addition of structural rules.

Much of the interest in using categorial grammars for
linguistic research derives from the possibilities they offer for
characterizing a flexible notion of constituency. This has been found
particularly useful in the development of theories of coordination,
and incremental interpretation.
For example, assuming standard lexical type assignments, the following
right node raised sentence cannot be derived in {\sf AB}, but does
receive a derivation in a system which includes {\em R3}, with
each conjunct assigned the type indicated.
\begin{examples}
\item \label{ex:rnr}
   {} [John resents $_{S/NP}$] and [Peter envies $_{S/NP}$] Mary
\end{examples}

A calculus which includes  composition, {\em R2}, will allow  a
function to apply to an unsaturated argument, and it is this property
which allows Ades and Steedman \cite{steedman82} to treat long
distance dependencies, and motivates much of Steedman's later work on
incremental interpretation.

 Dowty \cite{dowty88} uses the combination of composition, {\em R2}
 and lifting,  {\em R4},
 to derive examples of non-constituent
coordination such as  {\em John gave mary a book and Susan a record}.

We can increase the power of {\sf L} by adding the
structural transformations {\em Permutation}, {\em Contraction} and
{\em Expansion}, to derive the calculi  {\sf LP}, {\sf LPC}, {\sf LPE}
and {\sf LPCE}. The structural transformation  {\em Permutation}, which
removes the restrictions on the linear order of types, allows us
to go beyond the purely concatenative derivations of {\sf L}.
This allows us to deal with sentences exhibiting non-standard constituent
order. For example, Moortgat suggests using permutation for dealing
with {\em heavy NP-shift} in examples similar to the following
\cite{moortgat88}:

\begin{examples}
\item \label{ex:comics}
  John gave [to his nephew $_{PP}$]
   [all the old comic books which he'd collected in his troubled
   adolescence $_{NP}$].
\end{examples}

In (\ref{ex:comics}), the bracketed constituents can be ``rearranged'' via
permutation so that a derivation is possible that employs the standard
type ((NP$\backslash$ S)/PP)/NP for the ditransitive verb {\em
  gave}.

 In  {\sf L}, while it is possible to specify a type
missing an argument on
its left or right periphery, it is not possible to specify a type
missing an argument ``somewhere in the middle'',  making it
impossible to deal with non-peripheral extraction. However,
as Morrill et al show, permutation provides the additional power
necessary to account for this phenomenon \cite{morrilletal90}.

In addition to permutation, there are also linguistic examples which
motivate  contraction (e.g. gapping, \cite{moortgat88}) and expansion
(e.g. right dislocation, \cite{moortgat88}). However it is universally
recognized that a system employing the unrestricted use of structural
transformations would be far too powerful for any useful
linguistic application, since it would allow arbitrary word order
variation, copying and deletion. For this reason, a goal of current
research is to build a system in which the resource  freedom of the
more powerful calculi can be exploited when required,  while the basic
resource  sensitivity of {\sf L} is retained in the general case. One
such approach is to employ structural modalities
\cite{morrilletal90}, which are operators that explicitly mark those
types which are permitted to be manipulated by specific structural
transformations.\footnote{Systems which allow the selective use of structural
transformations may be implemented in the general framework presented
here, although we do not address this issue.}

\section{A framework for Categorial Deduction}
\label{sec:Framework}

In this section we describe the theorem proving framework for
categorial deduction. We start by setting up basic ideas of categorial
logic, giving formal definitions of the core logical language. Then we
move on to the  theorem proving strategy, introducing the LKE approach
\cite{dagostino94} and the algebraic apparatus used to characterise
different calculi.

\subsection{The core syntax}

We assume that there is a finite set of atomic grammatical categories
which will be represented by special symbols: NP for
noun phrases, S for sentences, etc. So, the set of
well-formed categories can be defined as below.

\DEF{The set of well-formed categories, \cC is the smallest set which
  contains every basic category and which is closed under the following
  rule: \bit
\item[(i)] If X \EL \cC and Y \EL \cC, then X/Y, X\BS Y and X \BU Y \EL \cC
\eit
}{def:wff}

Our purpose in this section is to define a prcedure
which will enable us to verify, given an entailment relation \EN,
whether or not such a relation holds for the logic
being considered.

Many proof procedures for classical logic have been proposed: natural
deduction, Gentzen's sequents, analytic (Smullyan style) tableaux,
etc. Among these, methods which conform to the sub-formula principle
are particularly interesting, as far as automation is concerned. See
\cite{fitting90a} for a survey. Most of these methods, along with
proof methods developed for resource logics, such as Girard's proof
nets (a variant of Bibel's connection method), can be used for
categorial logic. Leslie \cite{leslie90} presents and compares some
categorial versions of these procedures for the standard Lambek
calculus L, taking into account complexity and proof presentation
issues. Although tableau systems are not discussed in
\cite{leslie90}, a close relative, the cut-free sequent calculus is
presented as being the one which represents the best compromise
between implementability and display of the proof.

Smullyan style tableau systems, however, have been shown to be inherently
inefficient \cite{dagostino94a}. They cannot even simulate
truth-tables in polynomial time. The main reason for this is the fact
that many of the Smullyan tableau expansion rules cause the proof tree to
branch, thus increasing the complexity of the search. Moreover,
keeping track of the structure of the derivations represents an extra
source of complexity, which in most categorial parsers
\cite{moortgat92,morrill95} is reflected in expensive unification
algorithms employed for dealing with substructural implication. In
order to cope with efficiency and generality, we have chosen the LKE
system \cite{dagostino94} as the proof theoretic basis of our
approach\footnote{For standard propositional logic, it has been shown
  that LKE can simulate standard tableau in polynomial time, but the
  converse is not true.}. LKE is an analytic (its derivations exhibit
the sub-formula property) method of proof by refutation\footnote{A
  formula is proved by building a counter-model for its negation.}
which has only one branching rule. In addition, its formulae are
labelled according to a labelling algebra which will determine the
closure conditions for the proof trees\footnote{See section
  \ref{sec:Comparison} for a discussion of how LKE, unlike proof nets
  or standard tableaux, enables us to reduce the computational cost of
  label unification.}. In what follows, we shall concentrate on
explaining our version of the system, the heuristics that we have found
useful for dealing with particularities of the calculi covered, and
the relevant results for these calculi. The usual completeness and
soundness results (with respect to the algebraic semantic provided)
are already given in \cite{dagostino94}, so we will not discuss them
here.

We have mentioned that the condition for a branch to be considered
closed in a standard
tableau is that both a formula and its negation occur on it. The
calculus defined above presents no negation, though. So, we have to
appeal to some extrinsic mechanism  to express
contradiction. In Smullyan's original formulation, the formulae
occurring in a derivation were all preceded by {\em signs}: T or F.
For instance, assume that we want to prove A \IM A in classical logic.
We start by saying that the formula is false, prefixing it by F, and
try to find a refutation for F A \IM A. For this to be the case both
T A (the antecedent) and F  A (the consequent) have to be the case,
yielding a contradiction. In classical logic we can interpret T and F
as assertion and denial respectively, and so we can incorporate F into
the language as negation,
obtaining uniform notation by eliminating the need for signed
formulae. In our approach, since negation is not defined in the
language, we shall make use of signed formulae as proof theoretic
devices. T and F will be used to indicate whether or not a certain string
available for combination to produce a new one.

\subsection{The generalised parsing strategy}

If we had restricted the system to dealing with signed formulae, we
would have a proof procedure for an implicational fragment of standard
propositional logic enriched with backwards implication and
conjunction. However, we have seen that the Lambek calculus does not
exhibit any of the structural properties of standard logic, and that
different calculi may be obtained by varying structural
transformations. Therefore, we need a mechanism for keeping track of
the structure of our proofs. This mechanism is provided by labelling
each formula in the derivation with {\em information
  tokens}\footnote{See \cite{gabbay94} for a proof theoretic
  motivated, LDS approach, and \cite{barwiseetal95} for an approach
  based on a finer-grained, semantically motivated information
  structure.}.

Labels will act not only as mechanisms for encoding the structure of
the proof, from a proof-theoretic perspective, but will also serve as
means to propagate semantic information through the derivation. A
label can be seen as an information token supporting the information
conveyed by the signalled formula it labels. Tokens may convey
different degrees of informativeness, so we shall assume that they are
ordered by an anti-symmetric, reflexive and transitive relation, \PO,
so that an expression like x \PO y asserts that y is at least as
informative as x (i.e. verifies at least as many sentences as x). We
also assume that this semantic relation, ``verifies'', is closed under
deductibility.

It is natural to suppose that, as well as categories, information
tokens can be composed. We have seen that a type S/NP
can combine with a type NP to produce an S. If we assume that there
are tokens x and y verifying respectively S/NP and NP, how would we
represent the token that verifies S? Firstly, we define a token
composition operation \CI. Then, we assume that, a priori, the order
in which the categories appear in the string matters. So, a minimal
information token verifying S would be x \CI y. As we shall see below,
the constraints we impose on \CI will ultimately determine which
inferences will be valid. For instance, if we assume that the order in
which the types occur is not relevant, then we may allow permutation
on the operands, so that x \CI y \PO y \CI x; if we assume that
contraction is a structural property of the calculus then the string
[S/NP, NP, NP] will also yield an S, since  y \CI y \PO y, etc.
Let's formalise these notions by defining an algebraic structure,
called {\em Information frame}.

\DEF{An Information Frame is a structures \cL = \OA\cP,\CI,\ID{},\PO\CA,
  where (i) \cP is a non-empty set of information tokens; (ii) \cP is
  a complete lattice under \PO; (iii) \CI is an order-preserving,
binary operation on \cP which satisfies continuity, i.e., for {\em every}
directed family \{z\emth{_{i}}\},
\emth{\bigsqcup} \{z\emth{_{i}} \CI x\} =  \emth{\bigsqcup} \{z\emth{_{i}}\}
\CI x
{\em and} \emth{\bigsqcup} \{x \CI z\emth{_{i}}\} =   x \CI
\emth{\bigsqcup}\{z\emth{_{i}}\}; and (iv) \ID{} is an identity element in \cP.
}{def:InformationFrame}

Combinations of types are accounted for in the labelling algebra by
the composition operator. Now, we need to define an algebraic
counterpart for syntactic composition, \BU, itself. When a formula
like S/NP \BU NP is verified by a token x, this is because its
components were available for combination, and consequently were
verified by some other tokens. Now, suppose S/NP was verified by a
token, say a. What would be the appropriate token for NP, such that
S/NP combined with NP would be verified by x? It certainly would not
be more informative than x. Moreover, if the expression S/NP \BU NP
were to stand for the composition of the (informational) meanings of
its components, then the label for NP would have to verify, when
combined with a, at most as much information as x. In order to express
this, we define the label for NP as being {\em the greatest y s.t.
x is at least as informative as a combined with y}. This token will be
represented by x \opRD a. In general, x \opRD y \ED \emth{\bigsqcup}
\{z\emth{\mid} y \CI z \PO x\}. An analogous operation, \opLD, can be defined
to cope with cases in which it is necessary to find the
appropriate  label for the first operand by reversing the
order of the tokens in the definition above. Some properties of
\opRD \cite{dagostino94}:
\begin{minipage}[b]{.48\linewidth}
\FOR{Property1}{y \CI (x \opRD y) & \PO & x}
\end{minipage}\hfill
\begin{minipage}[b]{.48\linewidth}
\FOR{Property2}{\ID{} & \PO & x \opRD x}
\end{minipage}
\begin{minipage}[b]{.48\linewidth}
\FOR{Property3}{(x \opRD y) \CI z & \PO & (x \CI z) \opRD y}
\end{minipage}\hfill
\begin{minipage}[b]{.48\linewidth}
\FOR{Property4}{(x \opRD y) \opRD z & \PO & x \opRD (y \CI z) }
\end{minipage} \\

Having set the basic elements of our proof-theoretic apparatus, we are
now able to define the components of a derivation as follows:
\DEF{Signed labelled formulae (SLF) are expressions of the form S Cat
  : L, where S \EL \{T,F\}, Cat \EL \cC and L \EL \cL}
{def:SignedFormula}
A derivation, or proof will be a tree structure built according to
certain syntactic rules. These rules will be called {\em expansion}
rules, since their application will invariably expand the tree
structure. There are three sorts of expansion rules: those which
expand the tree by generating two formulae from a single one occurring
previously in the derivation, those which expand the tree by combining
two formulae into a third one which is then added to the tree, and the
branching rule. The first kind of rule corresponds to what is called
\ALPHA{}-rule in Smullyan tableaux; these rules will be called
\ALPHA{}-rules here as well. The second and third kinds have no
equivalents in standard tableau systems. We shall refer to the
second kind as \SIGMA{}-rules, and to the branching rule as
\BETA{}-rule -- after Smullyan's, even though his branching rules are
different. \REF{fig:ExpansionRules} summarises the expansion rules to
be employed by the system. A deduction bar says that if the formula(e)
appearing above it occurs in the tree, then the formula(e) below it
should be added to the tableau. The rules are easily interpreted
according to the intuitions assigned above to signs, formulae and
information tokens. A rule like \ALPHA{(i)}, for example, says that if
A\BS B is not available for combination and x verifies such
information, then this is because there is an A available at some token a,
but the combination of a and x (notice that the order is relevant) does
not produce B.
\noindent
\begin{figure}[!h]
\centering\begin{tabular}{|l|l|l|l|l|l|l|}
\hline
\small{{\bf \ALPHA{}-rules}} & (i)  & (ii) &
                        (iii) &  \multicolumn{2}{c}{\small{{\bf
                            \BETA{}-rule}}} &\\
\hline
\underline{(\ALPHA{1})} & \underline{F A\BS B : x} &
\underline{F A/B : x}   & \underline{\raisebox{-.1ex}[.05ex][.6ex]{ T A
                                                     \BU B : x}} &
               \multicolumn{3}{c|}{}  \\
(\ALPHA{2}) & T A : a$^{*}$
           & T B : a$^{*}$         &   T A  : a$^{*}$  &
\multicolumn{3}{c|}{\emth{\overline{(\BETA{1}) \hspace{0.2cm} T A : x
\emth{\mid}
                                    (\BETA{2})\hspace{0.2cm} F A : x}}} \\
(\ALPHA{3}) & F B : a \CI x     & F A : x \CI a  &   T B : x \opRD a &
\multicolumn{3}{c|}{} \\
\hline
\small{{\bf \SIGMA{}-rules}} & (i) & (ii) & (iii) & (iv) & (v) & (vi) \\
\hline
(\SIGMA{1}) & T A\BS B : x & T A\BS B : x & T A/B : x & T A/B : x &
F A \BU B : x & F A \BU B : x \\
\underline{(\SIGMA{2})} & \underline{T A : y} & \underline{F B : y \CI x} &
\underline{F A: x \CI y} & \underline{T B : y} &
\underline{T A : y} & \underline{T B : y} \\
(\SIGMA{3}) & TB : y \CI x & F A : y & F B : y & T A : x \CI y & F B :
x \opRD y & F A : x \opLD y \\
\hline
\end{tabular}\\
\emth{\ast}:{\em a new label {\em a} (not occurring previously in the
derivation)
must be introduced.}
\caption{Tableau expansion rules}\label{fig:ExpansionRules}
\end{figure}
\noindent Given the expansion rules, the definition of the main data
structure to be manipulated by the theorem proving (parsing) algorithm
is straightforward: a derivation tree, \cT, is simply a binary tree
built from a set of given formulae by applying the rules. The next
step is to define the conditions for a tree to be regarded as
complete. Completion along with inconsistency are the notions upon
which the algorithm's termination depends. It can be readily seen on
\REF{fig:ExpansionRules} that for a finite set of formulae, the number
of times \ALPHA{} and \SIGMA{} rules can be applied increasing the
number of SLFs (nodes) in \cT$\;$ is finite. Unbounded application
of  \BETA{}, however, might expand the tree indefinitely. In order to
assure termination, applications of \BETA{} will be restricted to
sub-formulae of formulae in \cT.  These notions are formalised in
\RED{def:Completion}.
\DEF{(Tree Completion) Given \cT, a tree for a set of SLFs S, we say
  that a binary tree \cT * is a tableau for S if \cT * results from
  \cT$\;$ through the application of an expansion rule. A tableau \cT
  * is {\em linearly complete} if it satisfies the following
  conditions: (i) if \ALPHA{1} \EL \cT, then \ALPHA{2} and \ALPHA{3}
  \EL \cT; (ii) if \SIGMA{1} and \SIGMA{2} \EL \cT, then \SIGMA{3} \EL
  \cT. A tree \cT * is {\em complete} iff for every A \EL \cT * and
  every sub-formula A' of A, both F A' : x and T A' : x have been
  added to \cT* by an application of the
  \BETA{}-rule.}{def:Completion}
Now, the first step towards building a counter-model for the denial of
the formula to be proved is the search for a tree containing {\em potential}
contradictions. Whether or not a  potentially inconsistent tree is a
counter-model for the formula will depend ultimately upon the
constraints on the labelling algebra. This form of inconsistency is
defined below.
\DEF{(Branch and Tree Inconsistency) A branch  is {\em
    inconsistent} iff for some type X both T X and F X, labelled by
  any information token, occur in the branch. A tree is inconsistent
  iff its branches are all inconsistent.}{def:TreeInconsistency}
Given the definitions above, we are ready to define an algorithm for
expanding linearly the derivation tree. For efficiency reasons
non-branching rules will be exhaustively applied before we move on to
employing \BETA{}-rules. \RED{def:LinearCompletion} presents the basic
procedure for generating linear expansion for a branch.\footnote{The
  reader will notice that if we had allowed \SIGMA{2} formulae to
  search for \SIGMA{1} types for combination, in the same way that \SIGMA{1}s
  search for \SIGMA{2}s,
  then linear expansion would not terminate for some cases.
  Consider for example the infinite sequence of \SIGMA{} applications:
  T A / B: x, T B / A : y, T A : z, T B : y \CI z, T A : x \CI (y \CI
  z), T B : y \CI (x \CI (y \CI z)),...  A
  strategy to allow unrestricted \SIGMA{}-application  without running
  into non-terminating procedures, as well as
  other practical and computational issues is discussed in
  \cite{luz95c}.} The complete LKE algorithm, \RED{def:Expansion},
which uses the procedure below, will be presented after we have
discussed tableau closure from the information frame perspective.

\DEF{(Algorithm: Linear Completion) Given \cT, a LKE-tableau
  structure, we define the procedure: \vspace{-1.3ex}\textnormal{
\begin{tabbing}
\mbox{{\sf Linear-Completion(\cT)}} \\
99 \=     \kill
1 \> \aDO \= \cT{} \MI \ALPHA{}-completion(\cT) \\
2 \> \> formula \MI head[\cT] \\
3 \> \aWHILE  \= ( \NE completed(\cT) \= \aOR \= consistent(\cT) ) \\
4 \> \> \aDO \= \aIF \= \SIGMA{1}-type(formula) \\
5 \> \>\> \aTHEN \aDO \= formula\emth{_{aux}} \MI
search(\cT,\SIGMA{2}) \=
$\;\;$ \= \aCOMMENT{formula\emth{_{aux}} is a set of \SIGMA{2}-type slf's}\\
6 \> \>\>\> \aIF \= formula\emth{_{aux}} \emth{\not = \emptyset}
\> \aCOMMENT{\SIGMA{3}-set results from combining \SIGMA{1} to each
  \SIGMA{2} }\\
7\> \>\>\>\> \aTHEN \aDO \= \SIGMA{3}-set \MI
                 combine-labels(\SIGMA{1},formula\emth{_{aux}}) \\
8\> \>\>\>\>\> \SIGMA{3}-expansion \MI \ALPHA{}-completion(\SIGMA{3}-set) \\
9\> \>\>\>\>\> \cT \MI append(\cT,\SIGMA{3}-expansion) \\
10\> \> \aDO formula \MI next[\cT] \\
11\> \aRETURN \cT
\end{tabbing}
}}{def:LinearCompletion}

We have seen above that the labels are means to propagate information
about the formulae through the derivation tree. From a semantic
viewpoint, the calculi addressed in this paper are obtained by varying
the structure assigned to the set of formulae in the
derivation\footnote{For instance, resource sensitive logics such as
  linear logic are frequently characterised in terms of multisets to
  keep track of the ``use'' of formulae throughout the derivation.}.
Therefore, in order to verify whether a branch is closed for a
calculus one has to verify whether the information frame satisfies the
constraints which characterise the calculus. For instance, the
standard Lambek calculus \calcL{} does not allow any sort of structural
manipulation of formulae apart from associativity; \calcLP{} allows
formulae to be permuted; \calcLPE{} allows permutations and expansion
(i.e. if B can be proved from the sequent $\Delta$, A, $\Gamma$, then B can
be proved from $\Delta$, A, A, $\Gamma$); \calcLPC{} allows permutation and
contraction; etc. The definition below sets the algebraic counterparts
of these properties.

\DEF{An information frame is: (i) {\em associative} if x \CI (y \CI z)
  \PO (x \CI y) \CI z and (x \CI y) \CI z \PO x \CI (y \CI z); (ii)
  {\em commutative} if x \CI y \PO y \CI x; (iii) {\em contractive} if
  x \CI x \PO x; (iv) {\em expansive} if x \PO x \CI x; (v) {\em
    monotonic} if x \PO x \CI y, for all x, y, z \EL
  \cP.}{def:Constraints}

Now, we say that a branch is {\em closed} with respect to the
labelling algebra if it contains SLFs of the form T X : x and F X : y,
where x \PO y. Likewise, a tree is closed if it contains only closed
branches.  Checking for label closure will depend on the calculus
being used, and consists basically of reducing information token
expressions to a {\em normal form}, via properties
(\ref{Property1})--(\ref{Property4}), and then matching tokens and/or
variables that might have been introduced by applications of the
\BETA{}-rule according to the properties or combination of properties
(\RED{def:Constraints}) that characterise the calculus considered.  It
should be noticed that, in addition to the basic algorithm, heuristics
might be employed to account for specific linguistic aspects. Some
examples: (a) it could be assumed that all the bracketing for the
strings is to the right thus favouring an incremental approach; (b)
type reuse could be blocked at the level of the formulae, reducing the
the computational cost of searches for label closure, since most of
the calculi in the family covered by the system are resource
sensitive; (c) priority could be given to juxtaposed strings for
\SIGMA{}-rule application, etc. \RED{def:Expansion} gives the general
procedure for tableau expansion, abstracted from the heuristics
mentioned above.

\DEF{(Algorithm: LKE-completion) The complete tableau expansion for a
  LKE-tree \cT is given by the following procedure:
\textnormal{
\begin{pseudocode}{expansion(\cT)}
1 \> \aDO closure-flag \MI no \\
2 \> \aWHILE  \= \NE ( completed(\cT) \= \aOR \= closed-\cT = yes) \\
3 \>\> \aDO \= \cT \MI linear-completion(\cT) \\
4 \> \>\> \aIF  \= \NE consistent(\cT)  \aAND label-closure(\cT) \\
5 \> \>\>\> \aTHEN \aDO \= closure-flag \MI yes \\
6 \> \>\>\> \aELSE \aDO \= subformula \MI select-subformula(\cT) \\
7 \> \>\>\>\> subformula\emth{_{T}} \MI assign-label-T(subformula)\\
8 \> \>\>\>\> subformula\emth{_{F}} \MI assign-label-F(subformula) \\
9 \> \>\>\>\> \cT\emth{_{1}} \MI append(\cT,\{subformula\emth{_{T}}\}) \\
10\> \>\>\>\> \cT\emth{_{2}} \MI append(\cT,\{subformula\emth{_{F}}\}) \\
11\> \>\>\>\> \aIF \= ( expansion(\cT\emth{_{1}}) = yes \aAND
                        expansion(\cT\emth{_{2}}) = yes ) \\
12\> \>\>\>\>\> \aTHEN \aDO closure-flag \MI yes \\
13\> \aRETURN closure-flag
\end{pseudocode}}}{def:Expansion}

As it is, the algorithm defined above constitutes a semi-decision
procedure. This is due to the fact that even though the search space
for signed formulae is finite, the search space for the labels is
infinite. The labels introduced via \BETA{}-rules are in fact
universally quantified variables which must be instantiated during the
label unification step. This represents no problem if we are dealing
with theorems, i.e. trees which actually close. However, for completed
trees with an open branch, the task might not terminate.  In order to
overcome this problem and bind the unification procedure we restrict
label (variable) substitutions to the set of tokens occurring in the
derivation --- similarly to the way parameter instantiation is dealt
with by liberalized quantification rules for first-order logic
tableaux.  In practice, the strategy adopted to reduce label
complexity also employs the following refinements: (i) the tableau is
linearly expanded keeping track of the choices made when
\SIGMA{}-rules are applied (the options are kept in a stack); (ii)
once this first step is finished, if the tableau is still open, then
backtrack is performed until either the choices left over are
exhausted or closure is achieved; (iii) only then is the \BETA{}-rule
applied. This explains the role played by the heuristics mentioned
above.\footnote{Furthermore, as we will show in section \ref{sec:Comparison},
if associativity is allowed at the syntactic level then it is possible
to eliminate the branching rule for the class of calculi discussed
here.} We are now able to establish some results regarding the
reduction laws mentioned in section \ref{sec:Family}.

\begin{proposition1}[Reduction Laws]
  Let X, Y and Z be types, and \cL an information frame. The
  properties R1--R5 hold:
\end{proposition1}

\begin{proof}
  \label{prf:ReductionLaws}
The proofs are obtained by straightforward application of
\RED{def:LinearCompletion} and \RED{def:Expansion}. Below we
illustrate the method by proving (R1) and
(R2):
\bit
\item[(R1)] To prove right application we start by
  assuming that it is verified by the identity token \ID{}. From this we
  have: 1- T X/Y \BU Y : m, 2- F X : 1 \CI m = m. Then, we apply
  \ALPHA{(iii)} to 1 obtaining 3- T X/Y : n and 4- T Y : m \opRD n.
  The next step is to combine 3 and 4 via \SIGMA{(iv)} getting 5- T X
  : n \CI (m \opRD n). Now we have a potential closure caused by 5 and
  2. If we apply property (\ref{Property1}) to the label for 5 we find
  that n \CI (m \opRD n) \PO m, which satisfies the closure condition
  thus closing the tableau.
\item[(R2)] Let's prove left composition. As we did
  above, we start with: 1- T Z\BS Y \BU Y\BS X : m and 2- F Z\BS X : 1
  \CI m. Applying \ALPHA{(iii)} to 1 we get: 3- T Z\BS Y : a and 4- T
  Y\BS X : m \opRD a. Now, we may apply \ALPHA{(i)} to 2 and get: 5- T Z
  : b and 6- F X : b \CI m. Then, combining 3 and 5 via \SIGMA{(i)}:
  7- T Y : b \CI a. And finally 4 and 7 through the same rule: 8- T X
  : (b \CI a) \CI (m \opRD a). The closure condition for 8 and 6 is
  achieved as follows: \vspace{-1ex}
\[
\begin{array}{llll}
  (b \CI a) \CI (m \opRD a) & \PO b \CI (a \CI (m \opRD a)) &
  \textit{by associativity} \nonumber \\ & \PO b \CI m & \textit{by
    (\ref{Property1}) and \CI being order-preserving} &\nonumber
\end{array}
\]
\eit  \vspace{-4ex}\hfill\QED
\end{proof}

Even though \calcL{} does not enjoy finite axiomatizability, the results
above suggest that the calculus finds a natural characterization in LKE
for associative information frames. In particular,
the Division Rule ({\em R6}) can be regarded as \calcL{}'s
characteristic theorem, since it is not derivable in weaker calculi
such as \calcAB{}, \calcNL{}, and \calcF{}. If we do not allow associative
frames, we get \calcNL{}. Stronger calculi such as \calcLP{},
\calcLPE{}, \calcLPC{} and \calcLPCE{} \cite{moortgat93}
can be obtained for the same general
framework by assigning further properties to \CI in the labelling
algebra. Frames exhibiting combinations of monotonicity, expansivity,
commutativity and contraction allow us to characterise these
substructural calculi. Algebras that are both associative and
commutative describe \calcLP{}. Adding expansivity (weakening)
to \calcLP{} results in \calcLPE{}. Associativity,
commutativity and contraction describe \calcLPC{} frames. \calcLPCE{} is
obtained by combining the properties of \calcLPC{} and \calcLPE{}
algebras.

We end this section with a simple example requiring associativity:
show, in \calcL{}, that an NP ({\sf John}), combined with a type
(NP\BS S)/NP ({\sf
  likes}) yields S/NP, i.e a type which combined with a NP will result
in a sentence (Proof \ref{prf:JLBBHM}).
\noindent\begin{proof}Let's assume the following type--string
  correspondence:  NP for  {\sf John}, (NP\BS S)/NP for {\sf likes}.
  The expression we want to
  find a counter-model for is: 1- F NP \BU (NP\BS S)/NP \ENI{\! L}
  S/NP. Therefore, the following has to be
  proved: 2- T NP \BU (NP\BS S)/NP : m and 3-  F S/NP : m.
  We proceed by breaking 2 and 3 down
  via \ALPHA{(iii)}, obtaining:
4- T  NP  : a,
5- T  (NP\BS S)/NP  : m\opRD a,
6- T  NP  : b, and
7- F  S  : (m \CI b). \\
\noindent Now we start applying \SIGMA{}-rules (annotated on the
right-hand side of each line):\\
\noindent\begin{tabular}{llll}
8- T & NP\BS S & : (m  \opRD a) \CI b & 5,6 \SIGMA{(i)} \\
9- T & S & : a \CI ((m \opRD a) \CI b)  & 4,9 \SIGMA{(i)} \\
\end{tabular} \\
We have derived a potential inconsistency between 7 and 9. Turning
our attention to the information tokens, we verify closure for \calcL{}
as follows:
\begin{center} \begin{tabular}{lllll}
a \CI ((m \opRD a) \CI b) & \PO &  (a \CI (m \opRD a)) \CI b  & by
associativity  \\
& \PO &  m \CI a & by property (\ref{Property1})
\end{tabular}\end{center} \vspace{-3ex} \hfill \QED
\label{prf:JLBBHM}
\end{proof}

\subsection{Comparison with Existing Approaches}
\label{sec:Comparison}

Early implementations of CG parsing relied on cut-free Gentzen
sequents implemented via backward chaining mechanisms
\cite{moortgat88}. Apart from the fact that it lacks generality, since
implementing more powerful calculi would involve modifying the code in
order to accommodate new structural rules, this approach presents
several sources of inefficiency. The main ones are: the
generate-and-test strategy employed to cope with associativity, the
non-determinism in the branching rules and in rule application itself.
The impact of the latter form of non-determinism over efficiency can
be reduced by testing branches for {\em count invariance} prior to
their expansion and by performing sequent proof normalisation.
However, non-determinism due to splitting in the proof structure still
remains. As we move on to stronger logics and incorporate structural
modalities such problems tend to get even harder.

An improved attempt to deal uniformly with multiple calculi is
presented in \cite{moortgat92}. In that paper, the theorem prover
employed is based on proof nets, and the characterisation of different
calculi is taken care of by labelling the formulae. For substructural
calculi stronger than \calcL{}, much of the complexity (perhaps too
much) is shifted to the label unification procedures. A strategy for
improving such procedures by compiling labels into higher-order logic
programming clauses is presented in \cite{morrill95} for \calcNL{} and
\calcL{}. However, a comprehensive solution to the problem of binding
label unification, a problem which arises as we move from sequents
to labelled proof nets, has not been presented yet. Moreover, as
discussed in \cite{leslie90}, if we consider that the system is to be
used as a parser, as a tool for linguistic study, the proof net style
of derivation does not provide the clearest or most intuitive display of
the proofs.\footnote{Proof nets and sequent
  normalisation have also been employed to get around
  spurious ambiguity (i.e. multiple proof for the same
  sentence, with the same semantics). Our approach does not exhibit
  this problem.}

In our approach, the burden of parsing is not so concentrated in label
unification but is more evenly divided between the theorem prover and
the algebraic checker. This is mainly due to the fact that the system
allows for a controlled degree of non-determinism, present in the
\SIGMA{}-rules, which enables us to reduce the introduction of
variables in the labelling expressions to a minimum. We believe this
represents an improvement on previous attempts. Besides this,
controlling composition via bounded backtrack opens the possibility of
implementing heuristics reflecting linguistic and contextual
knowledge. In fact, we verify that, under the
appropriate application of rules, we are able to eliminate the
\BETA{}-rule for a class of theorems.

\begin{proposition1}[Elimination Theorem]
  All closed LKE-trees  derivable by the application of the set
  of rules \cR =
  \SE{\ALPHA{(i)},...,\ALPHA{(iii)},\SIGMA{(i)},...,\SIGMA{(vi)},\BETA{}} can
  be also derived from \cR $-$ \{\BETA{}\} + \{ assoc \}.
\end{proposition1}

The proof of this proposition can be done by defining an {\em abstract
  Gentzen relation}, proving a substitution lemma with respect to
the labelling algebra (as in \cite{dagostino94}), and showing that our
consequence relation is closed under the relevant Gentzen conditions
even if no \BETA{} rule is employed. The proof appeals to the fact that
no formula signed by F can occur in the sequents on the left-hand side
of the entailment relation, since the calculi presented here do not
have negation.\footnote{Of course, without the \BETA{} rule not all open
  trees generated will constitute downward saturated sets, since they
  might contain formulae which are not completely analysed.} We
believe that this result shows that, even though LKE label unification might
be computationally expensive for substructural logics in general,
the system seems to be well suited for categorial logics. We
refer the reader to \cite{luz95c} for a more comprehensive discussion
of these issues.

\section{Conclusions and Further Work}
\label{sec:Conclusion}

We have described a framework for the study of categorial logics with
different degrees of expressivity on a uniform basis, providing a tool
for testing the adequacy of different CGs to a variety of linguistic
phenomena. From a practical point of view, we have investigated
the effectiveness and generality issues of a parsing strategy for CG
opening an avenue for future developments. Moreover, we have pointed out
some strategies for improving on efficiency and for dealing with more
expressive languages, including structural modalities.

The architecture proposed seems promising. Its flexibility with
respect to the variety of logics it deals with, and its modularity
suggest some natural extensions to the present work.  Among them:
implementing a semantic module based on Curry-Howard correspondence
between type deduction and $\lambda$-terms, adding local control of
structural transformations (structural modalities) to the language,
increasing expressivity in the information frames for covering calculi
weaker than \calcL{} (e.g. Dependency Categorial Grammar
\cite{pickering93}), exploiting the derivational structure encoded in
the labels  to define heuristics for models of human
attachment preferences  etc.  Problems for further investigation might
include: the treatment of polymorphic types (by incorporating rules
for dealing with quantification analogous to Smullyan's \gD and \gG
rules \cite{fitting90a} \cite{smullyan68}), and complexity issues
regarding how the general architecture proposed here would behave
under more standard theorem proving methods.

\end{document}